\pdfoutput=1

\PassOptionsToPackage{usenames,dvipsnames}{xcolor}

\documentclass[11pt]{article}

\usepackage[preprint]{acl}

\usepackage{times}
\usepackage{latexsym}

\usepackage[T1]{fontenc}

\usepackage[utf8]{inputenc}

\usepackage{microtype}

\usepackage{inconsolata}

\usepackage{graphicx}
\usepackage{algorithm} % For pseudocode algorithm environment
\usepackage{algpseudocode}
\usepackage{algcompatible}
\usepackage{multirow} 
\usepackage{amsmath} 

\usepackage{booktabs}
    \usepackage{array,booktabs}

    \usepackage[capitalize]{cleveref}
    \crefname{section}{Sec.}{Secs.}
    \Crefname{section}{Section}{Sections}
    \Crefname{table}{Table}{Tables}
    \crefname{table}{Tab.}{Tabs.}
    \crefname{algocfline}{alg.}{algs.}
    \Crefname{algocfline}{Algorithm}{Algorithms}
    \crefname{algocf}{alg.}{algs.}
    \Crefname{algocf}{Algorithm}{Algorithms}

    \usepackage{enumitem}
    \usepackage{amssymb}

    \usepackage{soul}

    \newcommand{\ROne}[1]{{\color{BlueViolet}\textbf{{#1}}}}
    \newcommand{\RTwo}[1]{{\color{PineGreen}\textit{#1}}}

    \makeatletter
    \DeclareRobustCommand\onedot{\futurelet\@let@token\@onedot}
    \def\@onedot{\ifx\@let@token.\else.\null\fi\xspace}
    \def\eg{{e.g}\onedot}
    
    \makeatother

    \usepackage{mathtools}
    \usepackage{xspace}
    \usepackage{bm}
    \newcommand{\calP}{\mathcal{P}}
    \newcommand{\calR}{\mathcal{R}}
    \newcommand{\calS}{\mathcal{S}}

    \usepackage{microtype}
    \usepackage{inconsolata}
    \usepackage{graphicx}
    \usepackage{algorithm}
    \usepackage{algpseudocode}
    \usepackage{algcompatible}
    \usepackage{multirow} 
    \usepackage{amsmath} 
    
    \usepackage{booktabs}
    \usepackage{enumitem}
    \usepackage{amssymb}

    \usepackage{soul}

    \newcommand{\code}[1]{{\sethlcolor{light-gray}\hl{\texttt{#1}}}}
    \definecolor{light-gray}{gray}{0.95}

    \definecolor{colorq}{HTML}{4E95D9}
    \definecolor{colort}{HTML}{3B7D23}
\title{Piece of Table: A Divide-and-Conquer Approach\\
for Selecting Subtables in Table Question Answering}

\author{Wonjin~Lee$^*$ \; \; Kyumin~Kim$^*$ \; \; Sungjae~Lee \; \; Jihun~Lee \; \; Kwang~In~Kim  \\
  POSTECH \\
  \texttt{\{wonjin0403,kyuminkim,leeeesj,gmindflow,kimkin\}@postech.ac.kr} \\}

\begin{document}
\maketitle
{\renewcommand{\thefootnote}{*}\footnotetext{These authors contributed equally.}}
\begin{abstract}
Applying language models (LMs) to tables is challenging due to the inherent structural differences between two-dimensional tables and one-dimensional text for which the LMs were originally designed. Furthermore, when linearized tables are applied to LMs, the maximum token length constraints imposed by self-attention mechanisms make it difficult to comprehensively understand the context spread across large tables. To address these challenges, we present \emph{PieTa} (Piece of Table), a new framework for subtable-based question answering (QA). \emph{PieTa} operates through a multi-resolution iterative process: dividing tables into smaller windows, using LMs to select relevant cells within each window, and merging these cells to form a subtable. This approach enables the model to capture dependencies across multiple rows and columns while mitigating the limitations of long context inputs. Instantiated as a simple iterative subtable union algorithm, \emph{PieTa} achieves significantly improved performance over previous subtable-based QA approaches. 
\end{abstract}

\section{Introduction}
\label{s:intro}
Tables effectively mitigate data complexity by organizing information into curated rows and columns, making them more accessible and comprehensible than plain text. Consequently, various natural language processing %(NLP) 
tasks, including table question answering (QA), have developed to extract and interpret information from this structured format~\cite{ZXS17,ZS15,NHM22}.

However, applying (large) language models (LMs) to tables is challenging due to the inherent structural differences between tables and text. Text familiar to LMs is one-dimensional and is read in a linear sequence (\eg, in raster order), with contextual relationships confined within sentences. The meaning of a sentence can change dramatically if the token order is altered. In contrast, tables are inherently two-dimensional, requiring both horizontal and vertical interpretation. Unlike text, tables often lack a clear contextual flow between cells, and altering the order of columns or rows does not necessarily change their meaning~\cite{LHY24}.

\begin{figure}[t]
  \includegraphics[width=\columnwidth]{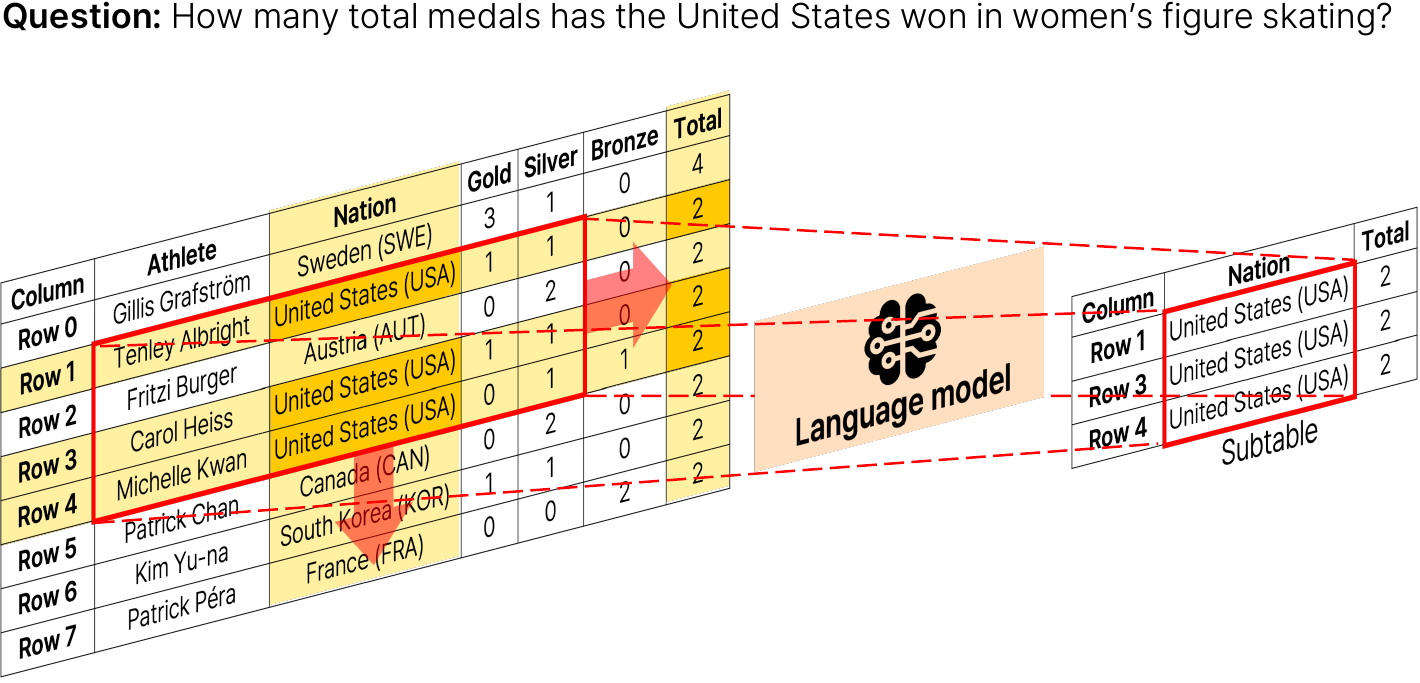}
  \caption{An overview of the proposed \emph{PieTa} (Piece of Table) framework. Starting with an input table and a question, our algorithm synthesizes a subtable by iteratively dividing the table into smaller windows, using language models to select relevant cells within these windows (forming intermediate subtables), and merging these until the final subtable is constructed. Code will be made publicly available upon acceptance.
  } 
  \label{f:overview}
\end{figure}

Recent advances in table QA have focused on accommodating these structural differences. For example, \citet{LCG22, JMH22} employed table QA-specific pre-training objectives guided by auxiliary SQL queries, while \citet{HPM20, YGU22, YNY20} explored the application of positional embeddings to linearized tables. A major limitation of these \emph{holistic} approaches lies in the computational constraint of transformer-based models, which typically restrict self-attention to 512 or 1,024 tokens, far from sufficient for handling large tables (see~\Cref{f:EM_BarCharts}). Consequently, these methods struggle with large tables or those with many irrelevant entries~\cite{PCA24}.

To overcome the limitations of holistic approaches, other methods focus on pre-articulating relevant information from tables. These methods first use LMs to translate the given questions into logical formats suited for tabular data retrieval, such as SQL or Python scripts~\citep{CXS23, NAS23}. Ideally, when well-generated, such code can retrieve relevant information effectively, enabling the subsequent QA process to bypass irrelevant distractions. However, enabling LMs to generate flexible programs that are versatile enough to accommodate the wide variations in tables remains a significant challenge~\citep{LWC23}.

Subtable selection methods aim to balance and mitigate the limitations of both holistic approaches and code generation methods. These methods first reduce the search space by extracting a relevant subtable, which is then fed to a table QA reader to construct the final answer.

For example, \citet{LBB23}'s \emph{inner table retriever (ITR)} selects rows and columns that align with the given question by calculating the similarity between the question and each individual row or column. This approach is computationally efficient and effective for answering \emph{simple} questions. However, it can struggle to capture dependencies across multiple rows and/or columns. For instance, when answering the question \code{Which was the next aircraft that came into service after Cessna 404 Titan?}, \emph{ITR} faces challenges because it must identify the \code{aircraft} that started the service after the \code{Cessna 404 Titan} without information on when it started the service.

\citet{YHY23}'s \emph{Dater} addresses this limitation by using LM in-context learning to jointly select multiple rows. While this improves upon independent row selection, it still inherits the limitation of one-dimensional LMs applied to two-dimensional tables, similar to the challenges encountered when generating answers directly from linearized tables. Furthermore, LMs generally struggle to understand and combine long context inputs~\cite{LLH24, WZL24}, a challenge particularly pronounced in table QA involving large tables.

In this paper, we present \emph{PieTa} (Piece of Table), a new framework for subtable-based QA. \emph{PieTa} operates through three main steps: %: divide, conquer, and combine. 
In the \emph{Divide} step, a given table is decomposed into smaller windows. In the subsequent \emph{Conquer} step, an LM selects cells from each window that match the given question (\Cref{f:overview}), constructing intermediate subtables. Finally, in the \emph{Combine} step, the extracted intermediate subtables from individual windows are merged to form a new table. This process is repeated, treating the resulting table as the new input, until the final subtable is constructed.

\emph{PieTa} effectively captures dependencies across multiple rows or columns while avoiding the challenge of long context inputs. This is achieved through a multi-resolution combination approach that enables the capture of long-term dependencies within the original table. Evaluated on the WikiTableQuestions (WikiTQ;~\citealp{ZS15}) and WikiSQL~\citep{ZXS17} datasets, \emph{PieTa} achieves significant improvements in QA accuracy and subtable selection performance.

\section{Related Work}
\label{s:relatedwork}

\noindent\textbf{Holistic table QA readers.\;\;} 
In table QA, a reader, often implemented as a language model (LM), generates answers to questions using either a complete table or a pre-processed subtable. For instance, the \emph{TaPEx} reader mimics an SQL execution engine by synthesizing answers from tables using SQL query pairs~\citep{LCG22}. It employs the \emph{BART} backbone model~\citep{LE20}, which is initially pre-trained on synthetic SQL queries and later fine-tuned to answer natural language questions.

Generating large-scale, high-quality table-SQL query pairs to articulate clear answers can be costly. \emph{OmniTab} simplifies this process by replacing complex SQL queries with plain text descriptions~\citep{JMH22}. It assumes that each table is accompanied by a textual description, which is used for pre-training before fine-tuning on table QA tasks.

These \emph{holistic} approaches inherit the limitation of LMs when applied to large tables, as typically only a small subset of cells in the table is relevant to a given question~\citep{PCA24}. Our approach efficiently identifies and selects these relevant cells, guiding the readers to generate more accurate answers.

\vspace{0.2cm}
\noindent\textbf{Guiding readers through subset selection.\;\;} 
Filtering irrelevant and redundant information from a table per question can help LM readers avoid distraction. \emph{Subtable selection} methods achieve this by constructing a tailored subtable for each question. \citet{LBB23} fine-tuned \emph{dense passage retriever} encoders~\citep{KOM20} to select relevant rows and columns. Their \emph{inner table retriever} (\emph{ITR}) employs supervised contrastive loss by labeling rows and columns containing \emph{answer} cells as positive. 
This technique was originally designed to ensure the input table size stays within the maximum token limit manageable by an LM. However, it does not explicitly construct a \emph{minimum} subtable that optimally balances precision and recall.
While extending this technique to filter irrelevant information 
is feasible, it cannot capture dependencies across multiple rows or columns (\Cref{s:intro}). In contrast, our method effectively reduces subtable size while retaining relevant information, leading to improved QA performance.

\emph{Dater} uses LM in-context learning to synthesize a subtable and sub-query pair from a given table and question~\cite{YHY23}. However, this approach requires extended prompts for the LM to fully comprehend the table structure and formulate an effective selection strategy, which becomes a severe burden for large tables. Generally, the performance of LMs tends to degrade when handling extremely long prompts~\cite{LLH24, WZL24}. 

\emph{TabSQLify} selects a subset of rows (with all columns) as input for LMs to generate SQL queries for subtable selection, significantly improving table QA performance~\cite{NR24}. However, similar to \emph{Dater}, it relies on LM in-context learning and faces the same limitation of requiring extended prompts. Our approach overcomes this by preemptively dividing large tables into smaller windows, allowing LMs to focus on extracting relevant cells within each window, alleviating the burden of processing long prompts.

\begin{figure*}[t]
  \includegraphics[width=\textwidth]{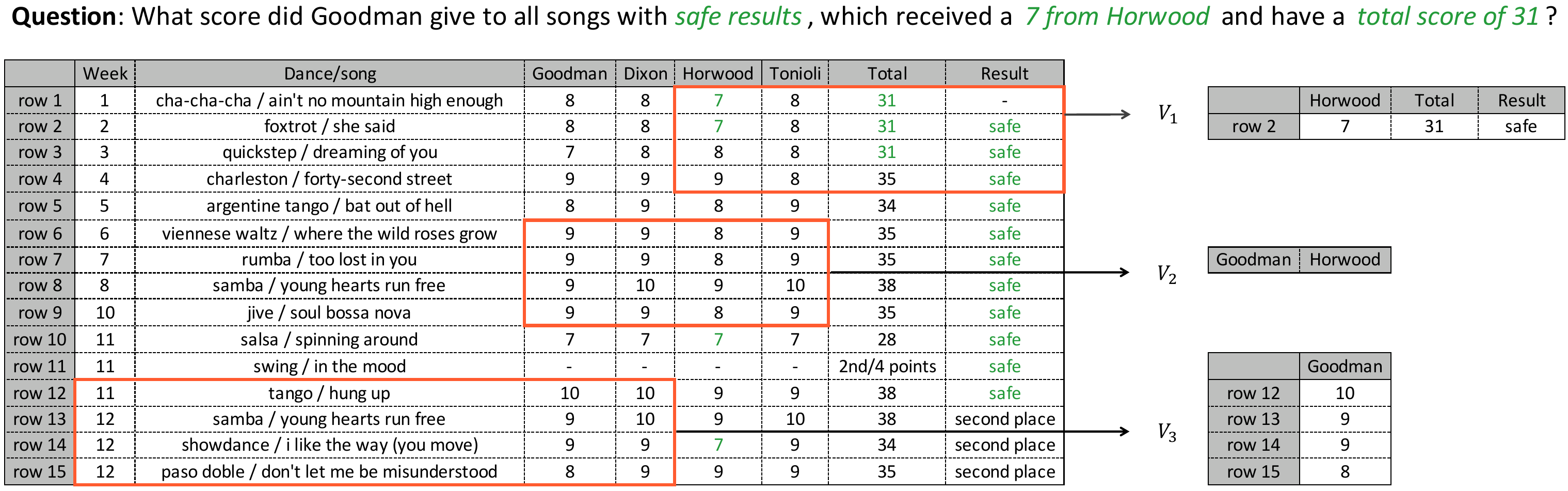}
  \caption{Examples of generated training data with an window size of $w=4$.
  The table and the question are sourced from the WikiSQL training dataset~\citep{ZXS17}. Three {\color{ForestGreen}{conditions}} and their corresponding {\color{ForestGreen}{cell values}} are color-coded in green. The three example input windows $\{W_i\}_{i=1}^3$, arranged in raster order, are highlighted with {\color{RedOrange}{orange boxes}}, while the corresponding training targets $\{V_i\}_{i=1}^3$ are shown on the right.
  }
  \label{f:data_generation}
\end{figure*}

\begin{figure}[t]
  \includegraphics[width=\columnwidth]{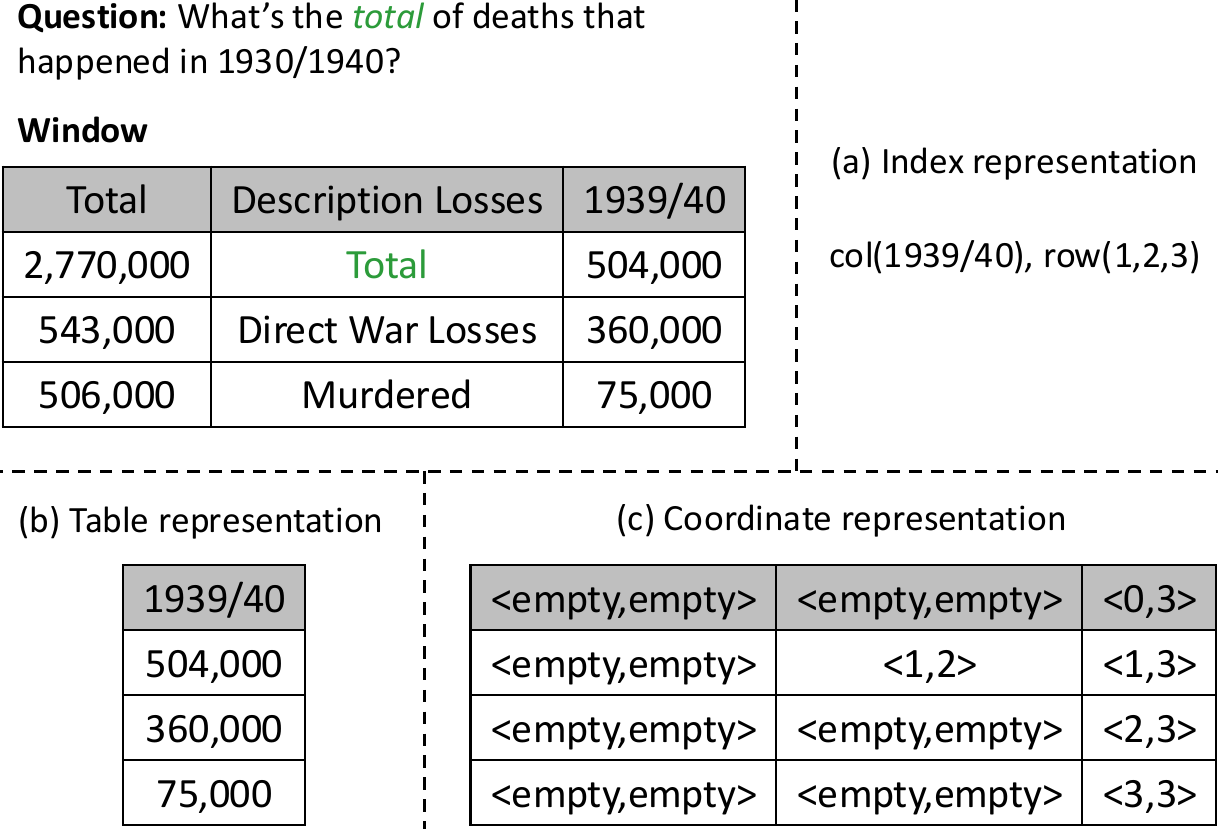}
  \caption{Examples of subtable representations ($w=3$). The information matching the condition \code{Total} is located in a cell rather than the column header. Both the index and table representations fail to select the cell \code{Total} because the column \code{Description Losses} is not selected. The proposed coordinate representation overcomes this limitation by preserving the input window structure and relevant cell contents.
  }
  \vspace{-0.05cm}
  \label{f:coordtoken}
\end{figure}

\section{Divide-and-Conquer Subtable Selection} 
\label{s:method}

The table question answering (QA) task requires generating an answer $A$ given a table $T$ and a human-understandable question $Q$. The answer $A$ can take various forms, such as a number, word, phrase, sentence, depending on the type of question and table. It is either extracted directly from $T$ or synthesized from the information within $T$ by the reader $\calR$. 

In subtable-based QA approaches, a selector $\calS$ first generates a subtable $T_s$ from $T$, 
which is then fed to the reader $\calR$ to generate the final answer:
\begin{align}
   A &= \calR(T_s, Q), \quad T_s = \calS(T, Q).\nonumber
\end{align}

Our study focuses on improving the selector $\calS$ and does not modify the readers. However, in principle, joint tuning of $\calR$ and $\calS$ is feasible.

\subsection{Subtable selection}
\label{s:main_al}
Our subtable selection algorithm iteratively applies three main steps. In the \emph{Divide} step, the input table $T$ is partitioned into small overlapping windows  $\{W_i\}_{i=1}^N$ of size $w\times w$ using a sliding window approach (\Cref{a:dividetable}). In the \emph{Conquer} step, the (subwindow) selector $\calS'$ extracts a subwindow $V_i$ within each window $W_i$, guided by the input question $Q$ and a prompt instruction $\calP$ (see \Cref{s:inputprompt_selector}):
\begin{align}
%\vspace{-0.2cm}
    V_i = \calS'(\calP(W_i, Q)).\nonumber
    %\label{e:subtable}
%\vspace{-0.2cm}
\end{align}
The \emph{Combine} step merges the generated subwindows $\{V_i\}_{i=1}^N$ into a new table $T^t$. Specifically, $T^t$ is obtained as the set union of all cells in the subwindows $\{V_i\}_{i=1}^N$. These steps are repeated, taking $T^t$ as the new input $T$ for the \emph{Divide} step in each iteration $t$, until the generated table $T^t$ is no longer updated. The complete subtable selection process is outlined in \Cref{a:subtable}. 

\begin{center}
\begin{algorithm}[t]
\caption{\textsc{SubtableSelection} ($\calS$).}
%\caption{Subtable selection module $\calS_\text{PieTa}$.}
\label{a:subtable}
\footnotesize
\begin{algorithmic}[1]
\STATE \textbf{Input:} Table $T$ and question $Q$ 
\STATE \textbf{Parameter:} Window size $w$
\STATE \textbf{Output:} Final subtable $T_s$

\STATE $t\gets 1$
\STATE $T^t \gets T$; $T^0 \gets \varnothing$
%\STATE Initialize \text{previous\_}$T_s \gets \emptyset$
\WHILE{$T^{t-1} \neq T^t$}
    \STATE $T^{t-1} \gets T^t$
    \STATE $\{W_i\}_{i=1}^N$ $\gets$ \textsc{DivideTable}($T^t$; $w$)\\ 
    \Comment{\emph{Divide} step; see~\Cref{a:dividetable}} 
    \STATE $T^t \gets \varnothing$
    \FOR{$W_i$ in $\{W_i\}_{i=1}^N$}
        \STATE $V_i^t \gets \calS'(\calP(W_i,Q))$ \Comment{\emph{Conquer} step }
    \STATE $T^t \gets T^t\cup V_i^t$ \Comment{\emph{Combine} step}
    \ENDFOR
    \STATE $t\gets t+1$
\ENDWHILE
\STATE $T_s \gets T^t$ 
\end{algorithmic}
\end{algorithm}
\end{center}

\subsection{Fine-tuning the selector}
\label{s:llmtuning}
 
We construct the selector $\calS'$ by fine-tuning the \emph{Llama3.1} model.\footnote{\texttt{Llama3.1-8B-Instruct}, available at\\ \quad\quad\quad\quad\quad\quad\url{https://www.llama.com/}.} 
In principle, however, our method is applicable to any LM-based selector.

\vspace{0.2cm}
\noindent\textbf{Data generation.\;\;} 
The training data are generated by sampling input windows (of size $w\times w$) from the SQUALL~\citep{SZB20} and WikiSQL~\citep{ZXS17} training datasets, which consist of tables, questions, and the associated \emph{target} subwindows (See~\Cref{s:experiments}).

We define \emph{condition columns} as columns containing cells that match the conditions specified in the question. Each cell in a condition column may or may not satisfy the given condition. Similarly, the column containing the exact cell values that the question seeks is referred to as the \emph{answer column}. For example, in the example question presented in \Cref{f:data_generation}, the columns \code{Horwood}, \code{Total}, and \code{Result} are the condition columns, while \code{Goodman} is an answer column.

For each training window $W_i$ sampled from a table $T$ with a corresponding question $Q$, the target subwindow $V_i$ is constructed by first selecting all condition and answer columns within $W_i$. Then, for these selected columns, only the rows that satisfy all the conditions present in $W_i$ are included.

\Cref{f:data_generation} illustrates the training data generation process. The first window $W_1$ contains three condition columns, \code{Horwood}, \code{Total} and \code{Result}. The target subwindow $V_1$ is then generated from the cells in these columns that jointly satisfy the conditions \code{7 from Horwood}, \code{total score of 31}, and \code{safe results}. The second window $W_2$ includes a condition column \code{Horwood} and an answer column \code{Goodman}. To create $V_2$, the subset of these columns satisfying the condition \code{7 from Horwood} is included. However, since no cells meet this condition, $V_2$ is constructed as an empty table containing only the column headers. The third window $W_3$ contains only an answer column. As no conditions are imposed, all cells from this column are included in $V_3$. 

This target window generation strategy enables the selector $\calS'$ to focus exclusively on the input windows $W$ when making predictions. See \Cref{s:details_train_data_sampling} for details.

\begin{figure*}[t]
  \includegraphics[width=\textwidth]{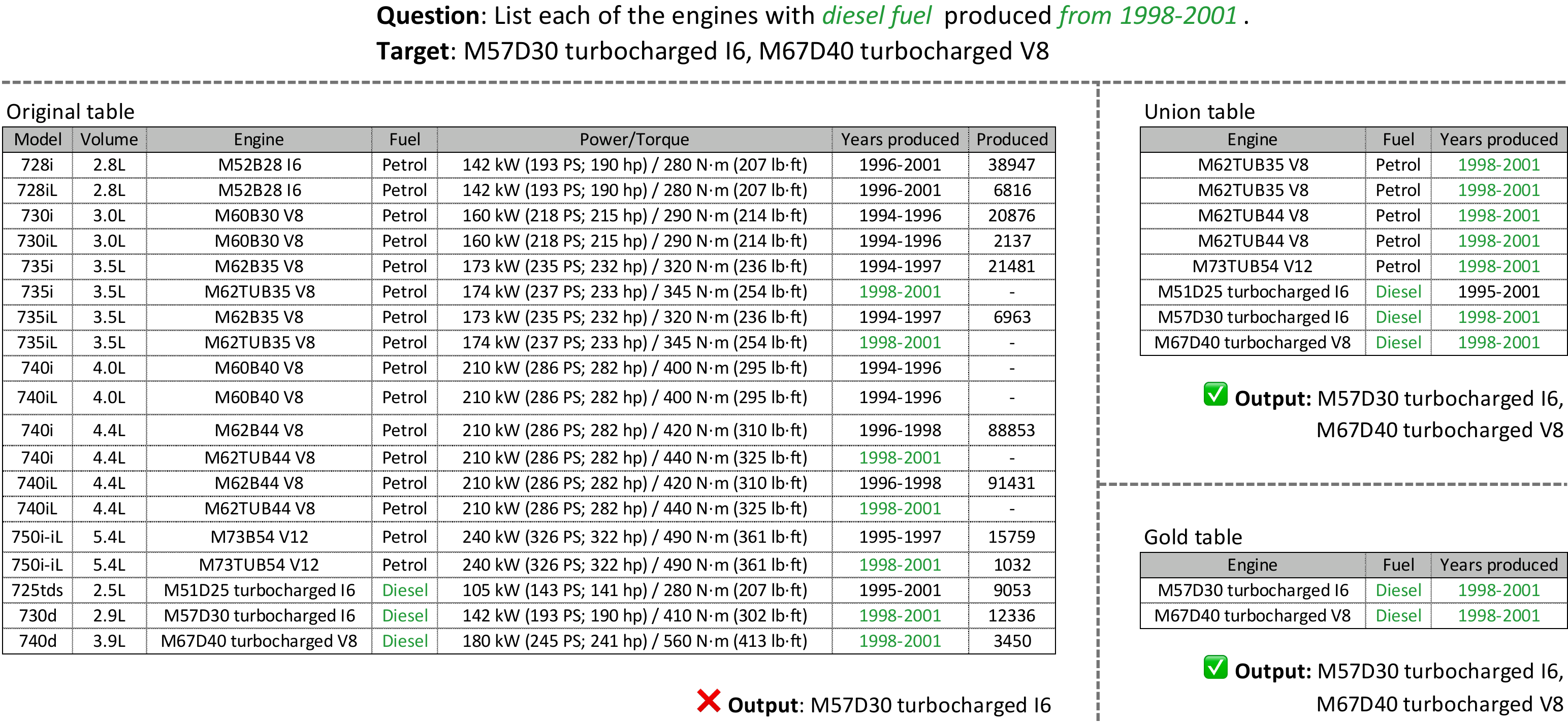}
  \caption{Examples of (sub)tables and the corresponding QA results. The original table has two condition columns (\code{Fuel} and \code{Years produced}) and one answer column \code{Engine}, which respectively match the conditions and expected outcomes presented in the question. Two {\color{ForestGreen}{conditions}} and the corresponding {\color{ForestGreen}{cell values}} are color-coded. The union table consists of rows satisfying at least one condition, while the underlying gold table consists of rows satisfying all conditions.
  Estimating the gold table is challenging as it requires a comprehensive understanding of the entire input table. The answers below the respective tables are the corresponding outputs of the \emph{TaPEx} reader. Feeding the original complete table leads to incorrect results, while both the gold and union table produce correct answers.
  }
  \label{f:example_wikitq}
\end{figure*}

\subsection{Representing subtables}
\label{s:representation}
Common representations for LMs in subtable generation include the \emph{index representation}, which identifies selected rows and columns using their indexes and headers, and the \emph{table representation}, which organizes selected cells into a structured table format. Both representations have inherent limitations that impact their effectiveness.

The index representation struggles when the information required to identify relevant content is embedded within cells rather than in row or column headers.
Because it relies solely on headers, it cannot directly capture cell content, making it difficult for selectors to learn how to align queries with embedded information.
For example, as shown in \Cref{f:coordtoken}(a), the column \code{Description Losses} is not explicitly mentioned in the question, while the content matching the condition \code{Total} appears within a cell. This mismatch prevents the LM from recognizing the column \code{Description Losses} as relevant.

In contrast, the table representation is prone to errors caused by the autoregressive nature of the token generation process. Since each token prediction depends on previous predictions, early errors in the sequence can propagate and compound, significantly reducing accuracy~\cite{BN24, SD10, BVJ15, LGZ16}. For instance, in \Cref{f:coordtoken}(b), if the LM fails to predict the column header \code{Description Losses}, all cells in that column are excluded.

To overcome these limitations, we propose the \emph{coordinate representation}, which preserves the input subwindow structure and uses \emph{coordinate tokens} (\code{<row\_index,column\_index>}) to indicate the positions of selected cells. Unselected cells are denoted by empty tokens (\code{<empty,empty>}).
This approach improves the ability of the model to identify relevant cells, even when their columns are initially overlooked, as demonstrated in \Cref{f:coordtoken}(c) with the successful identification of the cell \code{Total}.

\subsection{Discussion}
\label{s:discussion}
The crux of our approach lies in the multi-resolutional table formation technique, which builds upon table (sub)windows. In images, neighboring pixels often exhibit strong correlations, justifying the use of small windows \eg\ convolutional kernels in convolutional neural networks (CNNs) and attention windows in vision transformers. However, tables do not necessarily exhibit such spatial structures: For example, in the table shown in \Cref{f:example_wikitq}, neighboring cells like \code{1995-1997} and \code{1998-2001} are more similar within the \code{Years produced} column. In contrast, cells in the \code{Model 725tds} row lack a comparable relationship, \eg, \code{1995-2001} is not more closely related to \code{M51D25 turbocharged I6} than to \code{Diesel}. In this context, employing subwindows might initially appear to impose unnecessary restrictions.

Our union-based subwindow combination approach effectively overcomes such potential restrictions while ensuring focused answer generation by constructing moderately-sized final subtables: Suppose that the following question is presented for the table in \Cref{f:example_wikitq}: \code{List each of engines with diesel fuel produced from 1988-2001}. An ideal subtable for answering this question would intersect the cells meeting the conditions \code{diesel fuel} and \code{1998-2001}, along with their corresponding cell values in the \code{Engine} column, as illustrated in the bottom table in the right column of~\Cref{f:example_wikitq}. 

Directly identifying such a \emph{gold} table would require a comprehensive understanding of the entire table. Instead, our algorithm iteratively merges subwindows identified within each window. For example, it selects cells matching \code{Diesel} or \code{1998-2001} and combines them into a subtable, effectively taking their \textit{union} (as shown in the table in the upper right column of \Cref{f:example_wikitq}).

This union approach offers significant advantages. First, identifying relevant subwindows within small windows is much simpler than pinpointing the gold table, which requires intersecting conditions across the entire table: The selection module (LMs) can focus exclusively on the cells within each window, avoiding the challenges posed by long context inputs. Our training target generation process aligns with this design: Each target subwindow is constructed solely based on the conditions and outputs present within the corresponding input window. 

Furthermore, we empirically validated that the size of the resulting union subtables is substantially smaller than the original complete tables, facilitating subsequent answer generation steps. On average, the number of cells in the union tables was reduced to 13.91\% 
of the original table size. Despite being slightly larger than the gold tables, this approach delivers comparable performance for final table-based QA tasks. \Cref{t:motivexampleresult} shows an example of building subtables, demonstrating substantial improvements in final table QA performance.

\begin{table}[t]
  \centering
  \resizebox{\columnwidth}{!}
  {
    \begin{tabular}{ccccc}
        \toprule
        Dataset                  & Reader  & Original & Union & Gold  \\
        \midrule
WikiTQ                   & \emph{OmniTab} & 62.52    & 64.39& 67.79 \\
\midrule
\multirow{2}{*}{WikiSQL} & \emph{OmniTab} & 88.42    & 92.64 & 92.65 \\
                         & \emph{Llama3.1}  & 73.03    & 78.10 & 79.01 \\
        \bottomrule
    \end{tabular}
    }
    \caption{Table QA performance in exact match (EM; \%) across three different subtable types. The union tables we target achieve comparable performance to the gold tables while being significantly easier to construct.
    }
  \label{t:motivexampleresult}
\end{table}

\section{Experiments}
\label{s:experiments}

We evaluate the performance of the proposed \emph{PieTa} model using the WikiTQ~\cite{ZS15} and WikiSQL~\cite{ZXS17} datasets. Detailed descriptions of these datasets are provided in \Cref{s:expdetails}.
The condition and answer columns used to generate the target subwindow $V$ for a window $W$ are derived from the corresponding SQL queries. To annotate the target for each subwindow $V$, we used SQL queries from the SQUALL~\citep{SZB20} and WikiSQL datasets. SQUALL provides SQL annotations for 11,468 questions from WikiTQ.

The only hyperparameter of PieTa is the input window size $w$, which is fixed at $4$ across the entire set of experiments. The impact of varying window sizes is analyzed in \Cref{s:ablation}.

The evaluation is conducted from two perspectives. First, for table QA performance, subtables generated by \emph{PieTa} are fed into existing readers $\calR$ to obtain answers. We use \emph{OmniTab}~\cite{JMH22} and \emph{TaPEx}~\cite{LCG22}, fine-tuned on WikiTQ and WikiSQL, respectively. Additionally, to demonstrate \emph{PieTa}'s utility with generalist models not fine-tuned for table QA, we use \emph{GPT-3.5} and \emph{DP\&Agent}~\cite{LWC23} as readers (see~\cref{s:gptprompting} for details). Evaluation is based on the exact match (EM) score, which reflects the proportion of predictions that exactly match the ground truth. 

Second, for subtable selection, we evaluate the selection of condition and answer columns and rows using annotated labels from WikiSQL. Precision and recall are employed as metrics, where cells in the underlying gold table are \emph{positive} and those outside are \emph{negative}.

\begin{table}[t]
  \centering
  \setlength{\tabcolsep}{4.0pt}
  \resizebox{0.8\columnwidth}{!}
  {
    \begin{tabular}{lcc}
        \toprule
        Selector & Precision (P) & Recall (R)\\
        \cmidrule(r){1-3}
          \emph{ITR} & 13.45 & 97.84\\
          \emph{TabSQLify} & 75.47 & 87.36\\
          \cmidrule(r){1-3}
          \emph{Holistic LM} & 89.27 & 93.20\\
          \cmidrule(r){1-3}
          \cmidrule(r){1-3}
          \emph{PieTa} (1 iter.) & 19.22 & 99.65\\
          \emph{PieTa} (2 iter.) & 56.88 & 99.22\\
          \emph{PieTa} (3 iter.) & 60.13 & 99.15\\
          \emph{PieTa} (Final) & 60.88 & 99.12\\
        \bottomrule
    \end{tabular}
    }%
\caption{Performance of subtable selection algorithms for the subtable generation sub-task on WikiSQL, evaluated in precision (\%) and recall (\%). Hereafter, (P) and (R) denote precision and recall, respectively.
The last four rows highlight the effectiveness of our iterative subtable formation approach (see~\Cref{s:main_al}). The total number of iterations varies across instances, as \emph{PieTa} terminates once the subtable remains unchanged.
} 
\label{t:table_selection}
\end{table}

\subsection{Subtable selection results}

To evaluate \emph{PieTa} against existing approaches, we conducted experiments using \emph{ITR}~\cite{LBB23}, \emph{TabSQLify}~\cite{NR24}, and a simple baseline algorithm  that fine-tunes an LM to generate a subtable directly from the input table (referred to as \emph{Holistic LM}).\footnote{Both \emph{Dater} and \emph{TabSQLify} use in-context learning, whereas \emph{PieTa} employs fine-tuning. To the best of our knowledge, no existing fine-tuning approaches specifically target subtable selection. \emph{Holistic LM} can be considered a straightforward adaptation of \emph{Dater} for fine-tuning-based subtable selection.} Among these, \emph{TabSQLify}, \emph{PieTa}, and \emph{Holistic LM} use LM selectors. For \emph{PieTa} and \emph{Holistic LM}, we fine-tune \emph{Llama3.1} for one epoch (see \Cref{s:llmtuning}).
For \emph{TabSQLify}, we initially evaluated both \emph{Llama3.1} and \emph{GPT-3.5}, selecting the model that achieved the best performance (as \emph{GPT-3.5} was used in~\cite{NR24}).

\Cref{t:table_selection} summarizes the results. \emph{ITR}, which evaluates columns and rows independently, demonstrates low precision. In contrast, \emph{PieTa} effectively captures cross-column and row dependencies, resulting in precision improvement of 47.43\% over \emph{ITR}. 

Similar to \emph{PieTa}, \emph{TabSQLify} is capable of modeling joint dependencies. However, it relies on in-context learning for subtable selection, which necessitates extended prompts, thereby straining the LMs. \emph{TabSQLify}'s selection strategy first extracts a few rows (specifically, the first three rows, as in~\citealp{NR24}, which we follow in our experiments) and then determines condition and answer columns. A limitation of this approach is that if the initial row selection does not contain all required conditions, the algorithm may sample irrelevant columns, leading to suboptimal selections. Increasing the number of extracted rows could help address this issue, but it further burdens the LM selector due to longer context processing, ultimately resulting in plateauing performance that remains significantly lower than ours (see~\Cref{s:additionalresults}).

Another drawback of approaches that generate SQL queries, including \emph{TabSQLify}, is the risk of producing inoperable SQL code~\cite{XJX24}. We empirically observed that \emph{TabSQLify} generated unexecutable SQL code for 425 out of 15,878 pairs in the WikiSQL dataset and 83 out of 4,344 pairs in the WikiTQ dataset.

In subtable selection, achieving high recall is crucial, as missing relevant rows or columns can lead to incomplete information, ultimately degrading the performance of downstream table QA models. 
Among the compared subtable selection algorithms, only \emph{ITR} inherently produces high recall under its default settings provided by the authors. Other approaches lack a direct mechanism to balance precision and recall.
While \emph{PieTa} is not explicitly optimized for high recall, it consistently demonstrates a tendency to achieve higher recall compared to other approaches. \Cref{t:col_performance_by_gold_table_size} demonstrates that \emph{PieTa} maintains high recall across varying numbers of answer columns and rows.

\begin{table}
   \centering  
  \setlength{\tabcolsep}{3.5pt}
  \hspace*{-0.33cm}
  {
  \newcolumntype{Z}{ >{{}}r<{{}} }
  \newcolumntype{C}{ @{}>{{}}c<{{}}@{} }
  \newcolumntype{T}{ @{}>{{}}l<{{}} }
  \newcolumntype{W}{ @{}>{{}}l<{{}}@{} }
  {\small
    \begin{tabular}{ccc|ZCWcc}
        \toprule
        \multirow{2}{*}{\parbox{1.8cm}{\centering \# cond. \& ans. columns}}  
        & \multirow{2}{*}{(P)} 
        & \multirow{2}{*}{(R)} 
        & \multicolumn{3}{l}{\multirow{2}{*}{\# ans. rows}} 
        & \multirow{2}{*}{(P)} 
        & \multirow{2}{*}{(R)} \\ 
        & &  &  &  & \\
        \cmidrule(r){1-8}
        1 & 35.32& 100.00 & \;\;0  & -- &  5   & 60.83& 99.13\\
        2 & 66.68& \phantom{0}98.90& \;\;5  & -- & 10   & 65.14& 97.83\\
        3 & 48.37 & \phantom{0}99.54& \;\;10 & -- & 15  & 67.09& 98.82\\
        4 & 51.53& \phantom{0}99.69 & \;\;15 & -- & 20   & 77.54& 99.34\\
        5 & 42.76& \phantom{0}99.53& \;\;20 & -- & 270  & 72.06& 94.36\\
        \bottomrule
    \end{tabular}
    }
    }
  \caption{Subtable generation performance across varying numbers of condition and answer columns (left) and answer rows (right) on WikiSQL.}
  \label{t:col_performance_by_gold_table_size}
\end{table}

\begin{table}[h]
  \centering
  \resizebox{0.705\columnwidth}{!}
  {
    \begin{tabular}{llc}
        \toprule
                  Selector & Reader  & EM  \\
                  \midrule
                  \multirow{9}{*}{Holistic} & \emph{TaPEx}  & 56.54 \\
                  & \emph{OmniTab}  & 63.01 \\
                  & \emph{CABINET} & 60.00 \\
                  & \emph{GPT-3.5} & 60.75 \\
    & \emph{Binder} & 55.40 \\
    & \emph{StructGPT}  & 52.20 \\
    & \emph{Chain-of-Table} & 59.94 \\
    & \emph{ReAcTable} & 52.50 \\
    & \emph{DP\&Agent} & \RTwo{73.60}  \\
\midrule 
    \multirow{3}{*}{\emph{ITR}} & \emph{TaPEx} & 56.86 \\
     & \emph{OmniTab} & 63.47 \\
     & \emph{GPT-3.5} & 59.13 \\
    \midrule
    \multirow{4}{*}{\emph{Dater}} & \emph{TaPEx} & 53.73 \\
     & \emph{OmniTab} & 57.09  \\
     & \emph{GPT-3.5} & 57.78 \\
     & \emph{GPT-3.5$^\text{Dat}$} & 52.80 \\
    \midrule
    \multirow{4}{*}{\emph{TabSQLify}} & \emph{TaPEx} & 58.75\\
     & \emph{OmniTab} & 60.01  \\
     & \emph{GPT-3.5} &  59.21 \\
     & \emph{GPT-3.5$^\text{Tab}$} & 59.71 \\
    \midrule
    \multirow{4}{*}{\emph{PieTa}} & \emph{TaPEx} & 58.01\\
     & \emph{OmniTab} & 64.78\\
     & \emph{GPT-3.5} & 63.65\\
     & \emph{DP\&Agent} & \ROne{74.15}\\
    \bottomrule
    \end{tabular}}
    \caption{Table QA performance of subtable selector and reader combinations (EM; \%) on WikiTQ. The best and second-best results are highlighted in \ROne{bold} and \RTwo{italic} fonts, respectively.
    `Holistic' indicates that the original table is directly provided as input to the reader. For \emph{GPT-3.5$^\text{Dat}$} and \emph{GPT-3.5$^\text{Tab}$}, the GPT reader configurations from \emph{Dater} and \emph{TabSQLify} were used, respectively (see~\cref{s:gptprompting} for details).
    }
  \label{t:tableqa_wikitq}
  
\end{table}

\begin{table}[h]
  \centering
  \resizebox{0.65\columnwidth}{!}
  {
    \begin{tabular}{llc}
        \toprule
                  Selector & Reader  & EM  \\
        \midrule
    \multirow{5}{*}{Holistic} & \emph{TaPEx} & 89.31 \\
     & \emph{OmniTab} & 88.42 \\
     & \emph{CABINET} & 89.17 \\
     & \emph{GPT-3.5}  & 69.50 \\
    & \emph{StructGPT} & 65.60 \\
     \midrule
    \multirow{3}{*}{\emph{ITR}} & \emph{TaPEx} & 89.09 \\
     & \emph{OmniTab} & 89.71  \\
     & \emph{GPT-3.5}  & 67.60\\
     \midrule
    \multirow{4}{*}{\emph{TabSQLify}} & \emph{TaPEx} & 80.68 \\
     & \emph{OmniTab} & 81.05  \\
     & \emph{GPT-3.5}  & 69.58  \\
     & \emph{GPT-3.5$^\text{Tab}$}  & 70.25 \\
     \midrule
    \multirow{3}{*}{\emph{PieTa}} & \emph{TaPEx} & \ROne{91.83}\\
     & \emph{OmniTab} & \RTwo{91.17}\\
     & \emph{GPT-3.5}  & 75.49\\
        \bottomrule
    \end{tabular}}
    \caption{Table QA results on the WikiSQL dataset.
    }
  \label{t:tableqa_wikisql}
\end{table}

\subsection{Table QA results}
\Cref{t:tableqa_wikitq} presents the table QA performance of various subtable selection methods combined with different readers on the WikiTQ dataset. We also compared against state-of-the-art holistic table QA readers, including \emph{TaPEx}, \emph{OmniTab}, \emph{CABINET}~\cite{PCA24}, \emph{GPT-3.5}, \emph{Binder}~\cite{CXS23}, \emph{StructGPT}~\cite{JZD23}, \emph{Chain-of-Table}~\cite{WZL24}, \emph{ReAcTable}~\cite{ZHF24}, and \emph{DP\&Agent}~\cite{LWC23}.

Subtables generated by \emph{PieTa} improve reader performance by providing more focused and relevant information: Across most reader configurations, \emph{PieTa} consistently outperforms holistic readers and other subtable selection methods, showing significant improvements over original tables. 
A similar trend of improved performance is observed on the WikiSQL dataset (\Cref{t:tableqa_wikisql}).
The only exception is its combination with \emph{TaPEx}, where \emph{TabSQLify} yields slightly better performance on the WikiTQ dataset.

\begin{figure}[t]
  \includegraphics[width=\columnwidth]{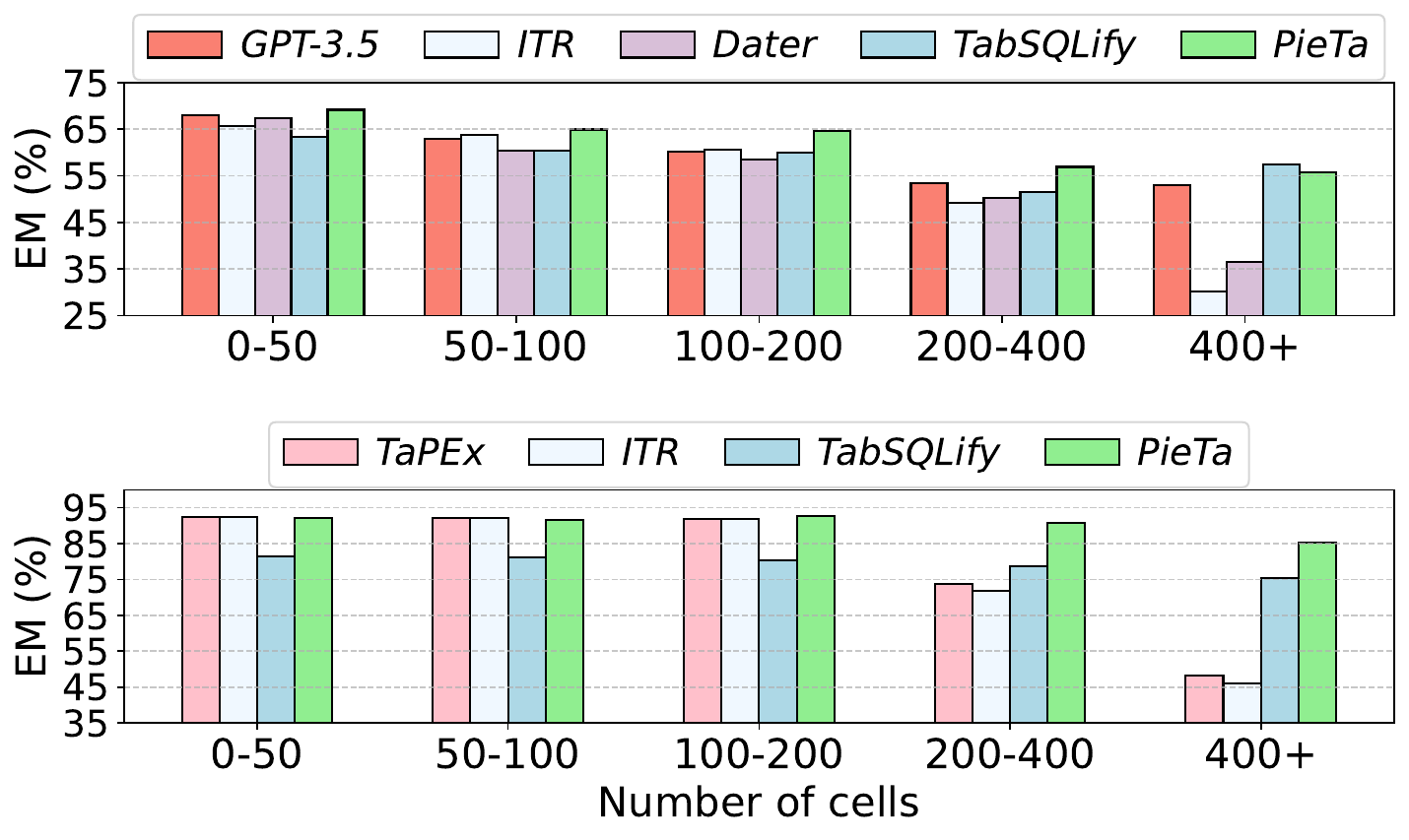}\vspace{-0.2cm}
  \caption{Table QA performance (EM; \%) across varying table sizes (number of cells) on WikiTQ (top) and WikiSQL (bottom).
  }
  \label{f:EM_BarCharts}
  \vspace{-0.2cm}
\end{figure}

Furthermore, \Cref{f:EM_BarCharts} demonstrates that \emph{PieTa} maintains strong table QA performance across varying input table sizes.

\subsection{Ablation study}
\label{s:ablation}
\noindent\textbf{Fine-tuning vs. in-context learning.\;\;}
We compare our fine-tuning approach for the LM selector with an alternative in-context learning method using a two-shot setup. As shown in \Cref{t:ablation}, in-context learning results in significantly lower performance compared to fine-tuning. This is primarily because LMs that are not specifically configured for subtable selection, often deviate from the intended task, \eg\ attempting to answer questions directly based on the subwindow content rather than generating relevant subtables. In contrast, fine-tuning aligns the selector's functionality with the subtable selection task and adapts the output format to a coordinate representation. This eliminates the need for lengthy prompts and improves overall performance.

\vspace{0.2cm}
\noindent\textbf{Subtable representation.\;\;}
\Cref{t:ablation} shows that the proposed coordinate-based subtable representation significantly outperforms existing index and table representations in table QA performance.

\vspace{0.2cm}
\noindent\textbf{Window sizes.\;\;}
The default window size for \emph{PieTa} is 4$\times$4. To evaluate its impact, we tested 3$\times$3, 5$\times$5, and 4$\times n$, where 4$\times n$ spans all columns (averaging approximately six columns; see~\Cref{s:expdetails}). As shown in \Cref{t:ablation}, both 4$\times$4 and 3$\times$3 achieve strong performance, while 5$\times$5 shows a slight decline, and 4$\times n$ degrades noticeably. This suggests that larger windows make it more challenging for the selector to identify cells that satisfy all conditions.

\begin{table}[t]
  \centering\resizebox{0.9\columnwidth}{!}{
  \setlength{\tabcolsep}{4.5pt}
    \begin{tabular}{ccc|c}
        \toprule
        Learning & Target repr. & Window size  & EM \\
        \midrule
        Fine-tune & Coord. & $4\times4$  & 90.11  \\
        \midrule

    In-context & Coord. & $4\times4$  & 87.60                                                                                                             \\
        Fine-tune & Index & $4\times4$  & 84.85 \\
        Fine-tune & Table & $4\times4$  & 89.44 \\
        Fine-tune & Coord. & $3\times 3$ & 90.11 \\
        Fine-tune & Coord. & $5\times 5$ & 89.71 \\
        Fine-tune & Coord. & $4\times n$   & 86.03 \\
        \bottomrule
    \end{tabular}}
  \caption{Table QA performance across variations of our algorithm on the WikiSQL validation set (\emph{TaPEx} reader). The first row represents our final method, integrating selector LM fine-tuning and a coordinate-based subtable representation. The results show that these design choices enhance performance individually and collectively.
  }
  \label{t:ablation}
  \vspace{-0.1cm}
\end{table}

\section{Conclusion}
We have explored the challenges of subtable selection in table question answering, focusing on the limitations of existing approaches in capturing joint dependencies across rows and columns, and managing lengthy token sequences in language models. Our algorithm bypasses these challenges by dividing input tables into small windows and independently constructing subtables therein. The resulting multi-resolution windowing strategy effectively combines subtables through simple union operations, streamlining subtable selection while preserving long-range relationships without imposing unnecessary operational assumptions. Evaluated on the WikiSQL and WikiTQ, our method demonstrated significant improvements over state-of-the-art methods. 

\section*{Limitations}
Our study focused on improving the subtable selection process independently of the reader that generates the final answer. This approach offers the advantage of flexibility, as the selection process can be easily combined with any type of reader. For instance, our algorithm can directly enhance the performance of readers that generate both natural language answers and SQL queries. 
However, in principle, tailoring the subtable selector specifically for the reader could further improve performance. Future research should explore the possibility of jointly optimizing the subtable selection process and the reader.

\section*{Acknowledgments}
This work was supported by Samsung Electronics Co., Ltd. (IO240508-09825-01), the National Research Foundation of Korea (NRF) grant (2021R1A2C2012195), and the Institute of Information \& Communications Technology Planning \& Evaluation (IITP) grant (RS-2019-II191906, Artificial Intelligence Graduate School Program (POSTECH)), funded by the Korea government (MSIT).
%(No.RS-2019-II191906, Artificial Intelligence Graduate School Program(POSTECH))

\bibliography{PieTaPaper}

\appendix

\section{Appendix}
\setcounter{figure}{0} \renewcommand{\thefigure}{A.\arabic{figure}}
\subsection{Subwindow sampling process}
\Cref{a:dividetable} details the window generation process from an input table. The \textsc{DivideTable} algorithm generates a set of subwindows $W = \{W_i\}_{i=1}^N$ by sliding a window of size $w\times w$ across the input table $T$ with a stride of 1.

\begin{center}
\begin{algorithm}[h]
\caption{\textsc{DivideTable}}
\label{a:dividetable}
\footnotesize
\begin{algorithmic}[1]
\STATE \textbf{Input:} Table $T$ with $R$ rows and $C$ columns
\STATE \textbf{Parameters:} Window size $w$
\STATE \textbf{Output:} Set of windows $W = \{W_i\}_{i=1}^N$

\STATE $W \gets \varnothing$
\FOR{$i=0$ \text{to} $R$ \textbf{by} 1}
    \FOR{$j=0$ \text{to} $C$ \textbf{by} 1}
        \IF{$i + w > R$}
            \State $i = R - w$
        \ENDIF
        \IF{$j + w > C$}
            \State $j = C - w$
        \ENDIF
        
        \STATE $W_i = T[i:i + w, j:j + w]$ \Comment{Window slicing}

        \STATE $W = W \cup \{W_i\}$
    \ENDFOR
\ENDFOR
\end{algorithmic}
\end{algorithm}
\end{center}

\subsection{Input prompt structure for selector $\calS$}
\label{s:inputprompt_selector}
\begin{figure*}[t]
\centering
  \includegraphics[width=1.5\columnwidth]{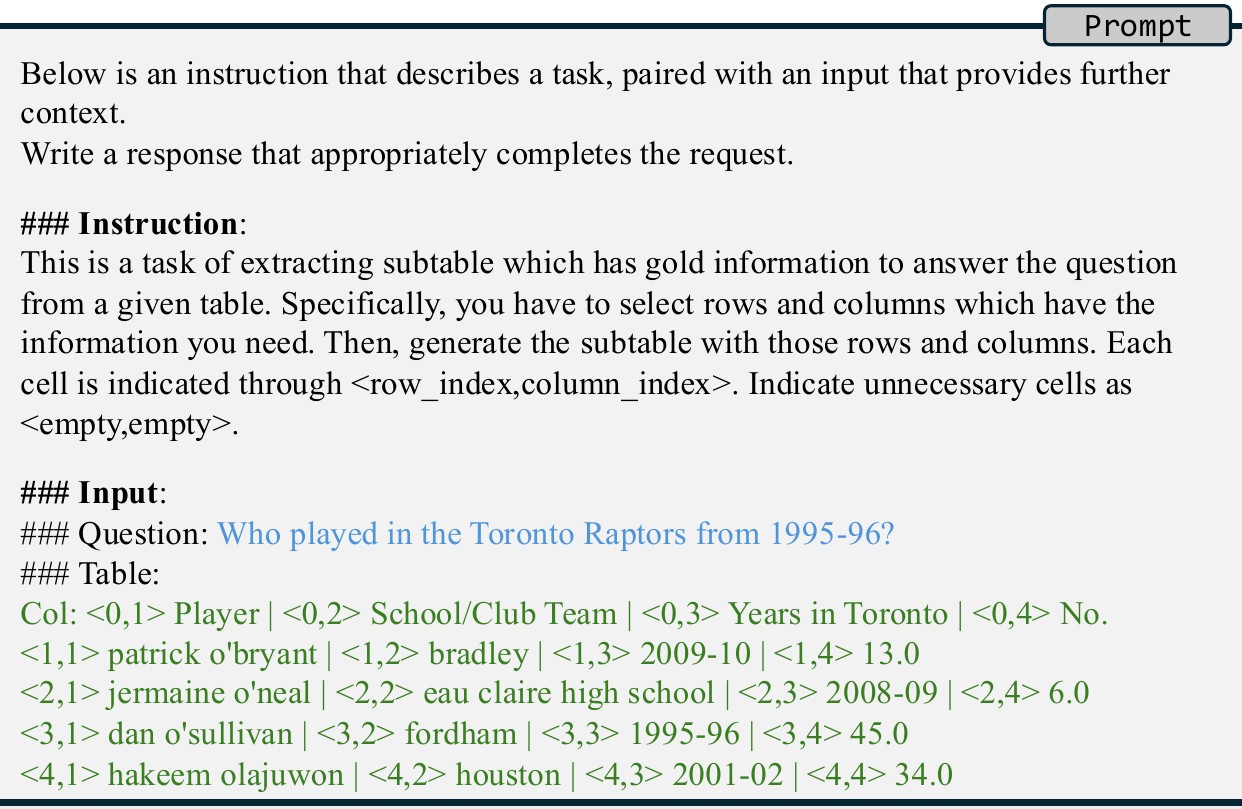}\\
  \includegraphics[width=1.5\columnwidth]{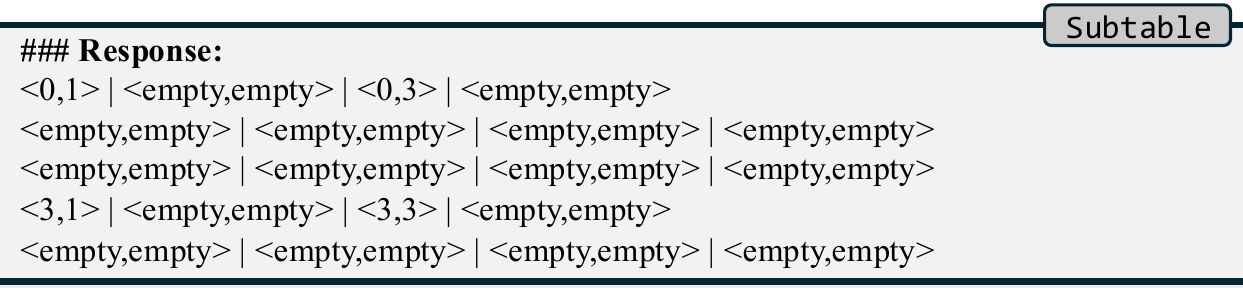}
  \caption{An example of a prompt. Given a question $Q$ and a window $W$, the prompt generator $\calP$ creates a textual prompt (top). This prompt is fed to the subwindow selector $\calS'$ to construct the output subwindow (bottom). The complete question $Q$ and input window $W$ are highlighted in {\color{colorq}{blue}} and {\color{colort}{green}}, respectively. In the instruction, the input and target windows are referred to as a \emph{table} and \emph{subtable}, respectively.
  }
  \label{f:prompt}
\end{figure*}

The input prompt for selector $\calS$ consists of both fixed and variable components, as shown in \Cref{f:prompt}. The variable components, highlighted in blue and green, represent the question and the table window, respectively. The fixed components remain unchanged across all tables and questions.

\subsection{Configurations of the \emph{GPT-3.5} reader}
\label{s:gptprompting}

\begin{figure*}[t]
\centering
  \includegraphics[width=1.5\columnwidth]{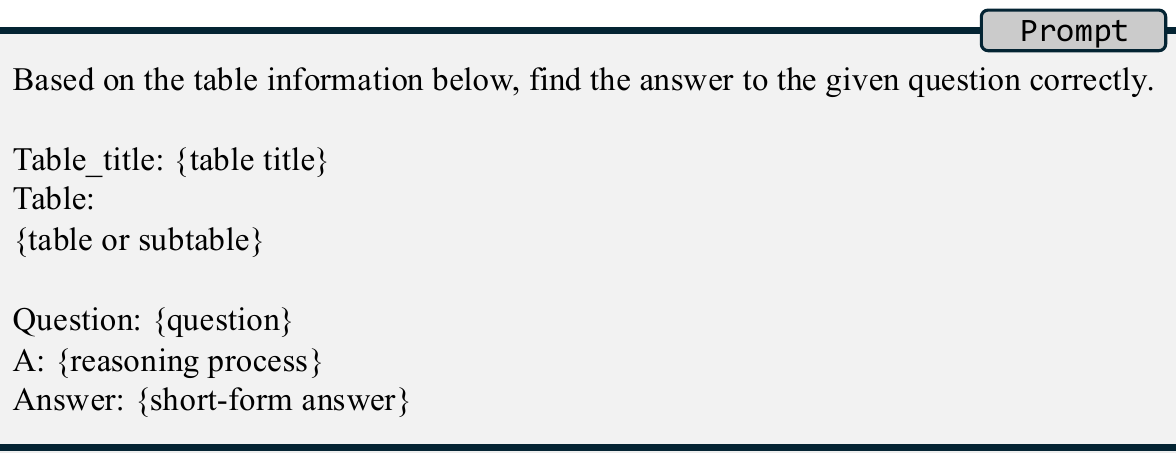}\\
  \includegraphics[width=1.5\columnwidth]{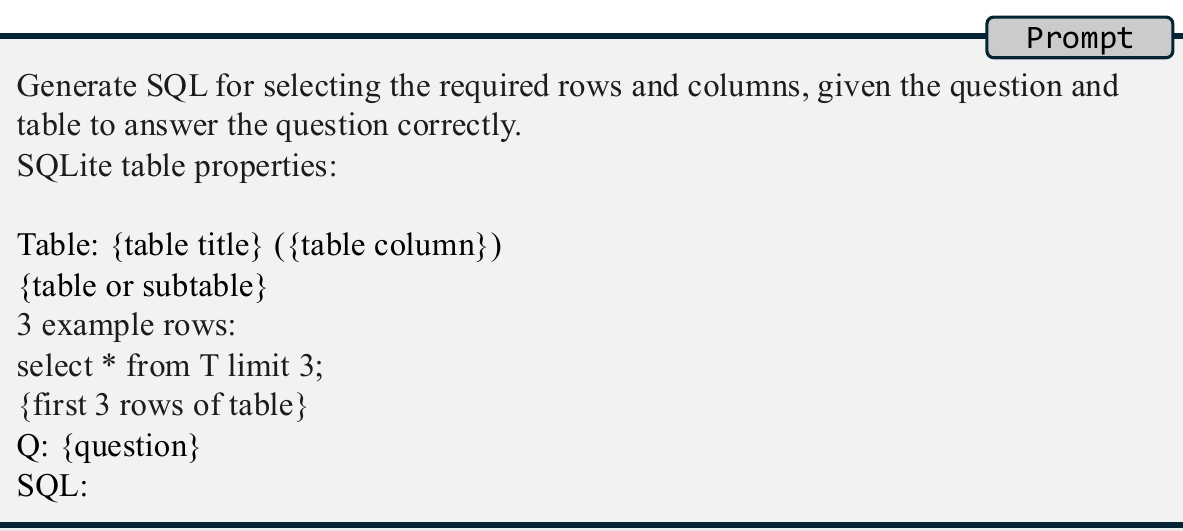}
    \caption{Template prompts for the holistic \emph{GPT-3.5} reader to generate textual answers (top) and SQL queries (bottom), adapted from~\cite{NR24}. Each \code{\{\}} placeholder in the prompt is replaced with the corresponding variable components.
  }
  \label{f:gpt_prompt}
\end{figure*}

For experiments with the \emph{GPT-3.5} reader in both the holistic setting and our approach (\Cref{t:tableqa_wikitq,t:tableqa_wikisql}), we allow \emph{GPT-3.5} to generate both SQL queries and textual output answers, following a similar strategy to \emph{DP\&Agent}~\cite{LWC23}. If the SQL query execution returns a single-cell result, the final answer is taken as the cell's contents. Otherwise, if the query produces multiple cells or is unexecutable, the textual output is used as the final answer. 

For both SQL and textual outputs, in-context learning is applied: 10-shot learning for SQL outputs following~\cite{NR24}, and two-shot learning for textual outputs, consistent with the experimental setup for holistic readers in~\cite{NR24}. The prompts used in these experiments are shown in \Cref{f:gpt_prompt}.

This setting differs from how the \emph{GPT-3.5} reader is used when combined with subtable selection in \emph{TabSQLify}~\cite{NR24} and \emph{Dater}~\cite{YHY23}. 
In \emph{TabSQLify}, the SQL query used to extract the subtable is incorporated into the reader prompt. In addition, when the extracted subtable contained only a single cell, it is omitted as input for the reader, and the cell itself was taken directly as the answer. In contrast, \emph{Dater}'s reader processes not only the subtable but also the outputs of a question decomposer, emphasizing question decomposition to effectively handle complex queries.
For a fair comparison, we also report results using these original \emph{GPT-3.5} configurations, denoted as \emph{GPT-3.5$^\text{Tab}$} and \emph{GPT-3.5$^\text{Dat}$} in \Cref{t:tableqa_wikitq,t:tableqa_wikisql}.

For \emph{DP\&Agent}, to prevent hallucination in \emph{GPT-3.5}, we provide the agent with the following additional guideline: \code{**Stay focused on the given question**: You must base your answer strictly on the given question. After the observation step, do not generate or infer new questions}.

\begin{figure*}[t]
  \includegraphics[width=\textwidth]{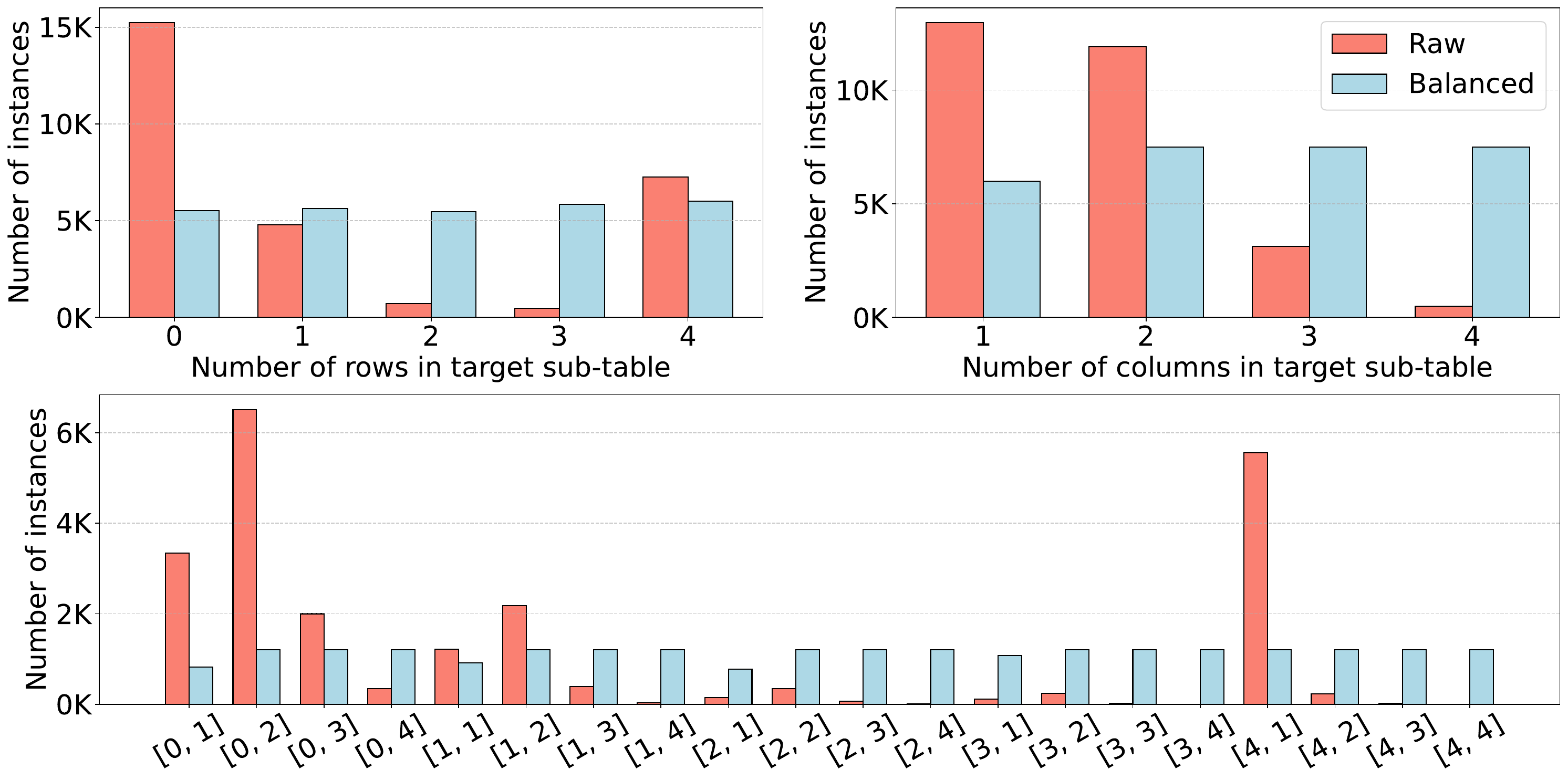}
  \caption{Histograms depicting the distribution of the generated $4 \times 4$ window training data for WikiSQL. The `raw' histogram represents the original, unfiltered data, while the `balanced' histogram shows the adjusted data used to mitigate class imbalance. In the bottom row, the x-axis represents [the number of answer rows, the number of condition \& answer columns]. Note that the y-axis scales vary across the histograms.
  }
  \label{f:raw_and_balanced}
\end{figure*}

\subsection{Details of experimental settings}
\label{s:expdetails}
\paragraph{Training settings.}
The selector is trained for a single epoch using an NVIDIA A100-80G GPU with the AdamW optimizer. The initial learning rate is set to $10^{-5}$. The batch size is 8, with 4 gradient accumulation steps. The training process takes approximately two hours for both WikiSQL and WikiTQ datasets.

\paragraph{Datasets.} Both WikiSQL and WikiTQ datasets are derived from tables extracted from Wikipedia, covering a wide range of domains.
WikiTQ includes operations such as superlatives, aggregations, and comparisons, with 4,344 test examples. On average, the test tables contain 6.3 columns and 25.8 rows, with some tables reaching up to 21 columns and 517 rows. WikiSQL features simpler questions involving aggregation, column selection, and conditions. It contains 15,878 test examples, with an average table size of 6.4 columns and 18.6 rows, and a maximum of 23 columns and 1,950 rows. Both datasets are widely used in table QA research, including~\cite{LBB23,NR24}.

Our experiments follow the experimental protocol of \citet{JMH22}, using a single run for each of the WikiTQ and WikiSQL datasets on clearly separated training, validation, and test sets. Our algorithm does not involve any random elements.

\paragraph{Table QA algorithms compared.}
For the experiments in \Cref{t:tableqa_wikitq,t:tableqa_wikisql}, we reproduced the results for \emph{OmniTab}, \emph{CABINET}, \emph{GPT-3.5}, \emph{ITR}, \emph{TabSQLify} and \emph{Dater} (excluding cases where the \emph{GPT-3.5$^\text{Dat}$} reader was used) using the code and services (for GPT) provided by the respective authors. However, for \emph{Dater} with \emph{GPT-3.5$^\text{Dat}$} and \emph{Binder} in \Cref{t:tableqa_wikitq}, the original studies relied on \emph{Codex}~\cite{CTJ21}, which was not available during our experiments. Instead, we report the results from~\cite{NR24} using \emph{GPT-3.5}. For other algorithms, we report the results as presented in their respective papers.

\paragraph{Data augmentation.}
We augmented the selector training data by labeling cells with the same value as target cells within the window as targets in the training dataset. This augmentation was applied to 8\% of the data in WikiTQ and 21\% in WikiSQL.

\subsection{Details of training data sampling}
\label{s:details_train_data_sampling}
In our preliminary experiments, we observed that randomly sampling training data (pairs of input $W_i$ and the corresponding target $V_i$ subwindows) often caused $\calS'$ to generate meaningless single-column target tables. This issue arises due to an imbalance in the distribution of target window sizes.

For a given input window $W$ of size $w\times w$, there are $w(w+1)$ possible target window sizes: $\{[1,\ 1],\ldots,[1,\  w],\ldots, [w,\ 1],\ldots,[w,\ w]\}$. Cases such as $[0,\ 0]$, $[1,\ 0]$, $[2,\ 0]$, \ldots, and $[w,\ 0]$ are excluded, as a table must contain at least one column.
Among the valid cases, single-column target windows dominate (\Cref{f:raw_and_balanced}), while cases where the entire input window $W$ becomes the target are rare. The latter scenario requires all columns in $W$ to be either condition or answer columns, with every cells satisfying the given question. To address this imbalance, we explicitly adjust the target window size distribution when sampling the training set.

To achieve a balanced distribution of the number of rows and columns in target subtables, we evenly match their occurrences. 
We first determine the number of columns $n\in\{1,2,\ldots,w\}$ in the target subtable and designate them as either condition or answer columns. The remaining $w-n$ columns are filled with non-condition, non-answer columns. Similarly, for the rows, we select $m$ rows that satisfy the condition and fill the remaining $w-m$ rows with cells that do not meet the condition.
By evenly sampling $n$ and $m$, we can generate a well-balanced dataset.

\subsection{Additional results}
\label{s:additionalresults}

\emph{TabSQLify} selects a subset of rows (specifically, the first three rows with all columns) as input for LMs to generate SQL queries for subtable selection. A key limitation of this approach is that if the initial rows do not contain all necessary conditions, the algorithm may sample irrelevant columns, leading to suboptimal selections. Increasing the number of extracted rows could help mitigate this issue, but it also burdens the LM selector with longer context processing, ultimately resulting in plateauing performance that remains significantly lower than ours (\Cref{t:tabsqlifyadditionalresults}).

\begin{table}[t]
  \centering\resizebox{0.8\columnwidth}{!}{
  \setlength{\tabcolsep}{4.5pt}
    \begin{tabular}{cc|c}
        \toprule
         \multicolumn{2}{c|}{Selector} & EM (\%)\\
         \midrule
         \multirow{4}{*}{\begin{tabular}{c}\emph{TabSQLify}\\ 
 (number of rows)\end{tabular}} 
         &3 (original)&59.71\\
         &10&60.52\\
         &20&61.58\\
         &Full&61.79\\
         \midrule
         \multicolumn{2}{c|}{\emph{PieTa}}&63.65\\
        \bottomrule
    \end{tabular}}
  \caption{Table QA performance of \emph{TabSQLify} (EM; \%) with varying numbers of rows, compared to our method, both using \emph{GPT-3.5} readers  (\emph{GPT-3.5$^\text{Tab}$} for \emph{TabSQLify}). See~\Cref{t:tableqa_wikitq} for further comparison.
  }
  \label{t:tabsqlifyadditionalresults}
  \vspace{-0.1cm}
\end{table}

\subsection{Potential risks}
\label{s:risks}
Since \emph{PieTa} relies on pre-trained language models (LMs) to extract relevant subtables, it may inherit and reflect any biases present in these models. Likewise, as \emph{PieTa} learns to identify subtables based on training data, it may inadvertently reinforce existing biases within the dataset.

While we do not foresee specific problematic scenarios, subtable selection inherently carries the risk of introducing unintended bias due to the selective sampling of table contents. Furthermore, in adversarial settings, subtable selection systems could be exploited to selectively present data that supports a particular agenda while omitting contradictory evidence.

\end{document}